\documentclass{article}
\usepackage{ijcai25}

\usepackage{times}
\usepackage{soul}
\usepackage{url}
\usepackage[hidelinks]{hyperref}
\usepackage[utf8]{inputenc}
\usepackage[small]{caption}
\usepackage{graphicx}
\usepackage{amsmath}
\usepackage{amsthm}
\usepackage{amsfonts}
\usepackage{booktabs}
\usepackage{algorithm}
\usepackage{algorithmic}
\usepackage[switch]{lineno}

\usepackage[caption=false,font=footnotesize,labelfont=rm,textfont=rm]{subfig}
\usepackage{multirow}

\usepackage{bbding}

\title{Boosting Few-Shot Open-Set Object Detection via Prompt Learning \\ and Robust Decision Boundary}

\author{
Zhaowei Wu$^1$
\and
Binyi Su$^{2,3}$\and
Qichuan Geng$^4$\and
Hua Zhang$^5$ \And
Zhong Zhou$^{1,6,}$\thanks{Corresponding author.}  \\
\affiliations
$^1$State Key Laboratory of Virtual Reality Technology and Systems, Beihang University, China\\
$^2$The School of Artificial Intelligence and Data Science, Hebei University of Technology, China\\
$^3$China Xiongan Group Digital City Technology Co., Ltd.\\
$^4$Information Engineering College, Capital Normal University, China\\
$^5$Institute of Information Engineering, Chinese Academy of Sciences\\
$^6$Zhongguancun Laboratory, Beijing, China
\emails
\{wuzhaowei, zz\}@buaa.edu.cn,
subinyi@hebut.edu.cn,
gengqichuan1989@cnu.edu.cn,
zhanghua@iie.ac.cn
}

\begin{document}

\maketitle
\begin{abstract}
Few-shot Open-set Object Detection (FOOD) poses a challenge in many open-world scenarios. It aims to train an open-set detector to detect known objects while rejecting unknowns with scarce training samples. Existing FOOD methods are subject to limited visual information, and often exhibit an ambiguous decision boundary between known and unknown classes. To address these limitations, we propose the first prompt-based few-shot open-set object detection framework, which exploits additional textual information and delves into constructing a robust decision boundary for unknown rejection. Specifically, as no available training data for unknown classes, we select pseudo-unknown samples with \textbf{A}ttribution-\textbf{G}radient based \textbf{P}seudo-unknown \textbf{M}ining (AGPM), which leverages the discrepancy in attribution gradients to quantify uncertainty. Subsequently, we propose \textbf{C}onditional \textbf{E}vidence \textbf{D}ecoupling (CED) to decouple and extract distinct knowledge from selected pseudo-unknown samples by eliminating opposing evidence. This optimization process can enhance the discrimination between known and unknown classes. To further regularize the model and form a robust decision boundary for unknown rejection, we introduce \textbf{A}bnormal \textbf{D}istribution \textbf{C}alibration (ADC) to calibrate the output probability distribution of local abnormal features in pseudo-unknown samples. Our method achieves superior performance over previous state-of-the-art approaches, improving the average recall of unknown class by 7.24\% across all shots in VOC10-5-5 dataset settings and 1.38\% in VOC-COCO dataset settings. Our source code is available at \href{https://gitee.com/VR\_NAVE/ced-food}{https://gitee.com/VR\_NAVE/ced-food}.
\end{abstract}

\section{Introduction}

Object detection ~\cite{ren2015faster,redmon2016you,lin2017focal} has achieved remarkable success, enabling downstream tasks by leveraging large-scale labeled training data under the close-set setting, where training and testing sets share the same classes. However, in real-world scenarios such as safe autonomous driving, training data typically follows a long-tail distribution and many unknown objects may lack clear definitions, which could potentially lead to serious safety risks with close-set detector. To address these limitations, Few-shot Open-set Object Detection (FOOD) \cite{su2024toward} enables the detection of known classes and rejection of unknowns with only few-shot close-set training data, which breaks the conventional assumption of identical class labels in training and testing. Despite its potential, this task is highly challenging due to limited training data and the absence of labels for unknown objects, resulting in weak generalization for unknown rejection and low recall rates.

\begin{figure}[!t]
\centering
\includegraphics[width=1\linewidth]{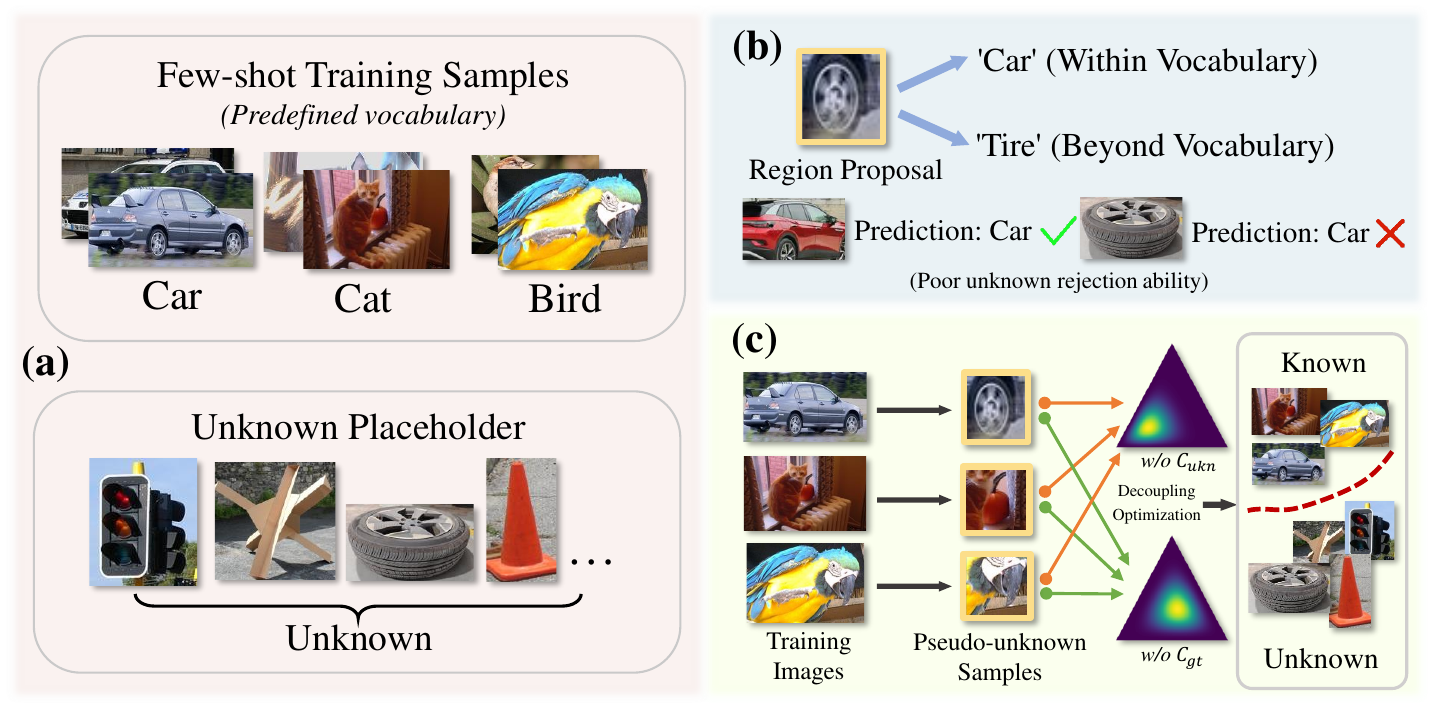}
\caption{The challenge of unknown classes in the image-text joint space and our solution. \textbf{(a)} There are still numerous unknown objects beyond the predefined vocabulary in real-world scenarios. \textbf{(b)} Our intuition is that region proposals with high uncertainty (yellow border) consist of features from both known and unknown classes. \textbf{(c)} 
Our method decouples and learns distinct information from pseudo-unknown samples to construct a discriminative decision boundary.
}
\vspace{-1em}
\label{fig:OS&OV}
\end{figure}

\begin{figure}
\centering
\includegraphics[width=1\linewidth]{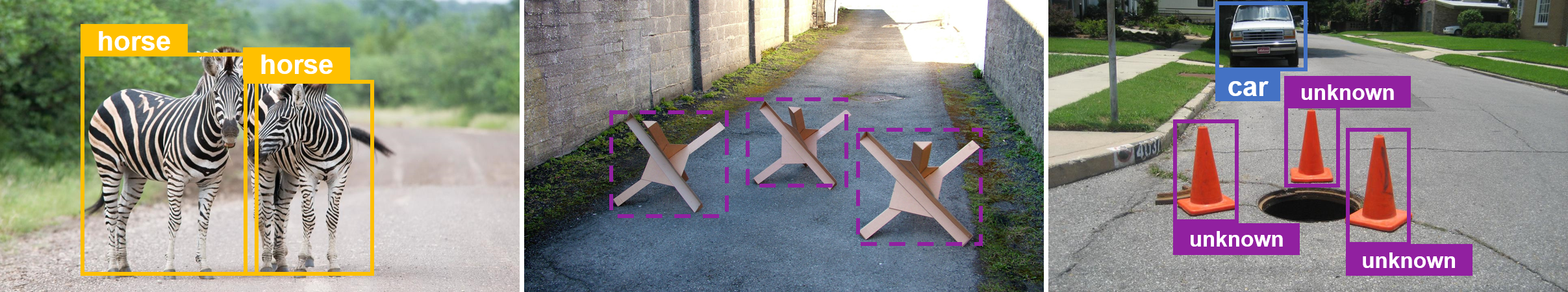}

\caption{The detector misidentifies the zebra as horses \textbf{(left)}. The detector misses the Czech hedgehog \textbf{(middle)}. The detector successfully detects the known and rejects the unknown \textbf{(right)}.}
\vspace{-1em}
\label{fig:false_missed_true}
\end{figure}

Existing FOOD methods have utilized weight
sparsification \cite{su2024toward} or moving weight average \cite{su2023hsic} to facilitate the generalization for unknown rejection. However, the decision boundary between known and unknown classes is still ambiguous due to limited visual information. This ambiguity often leads to the misclassification of unknown classes as known ones with high confidence, resulting in a relatively low recall rate. Recently, the vision-language models \cite{radford2021learning} introduce textual modality as complementary for downstream tasks through prompt learning \cite{zhou2022coop,zhou2022conditional}, shedding light on few-shot out-of-distribution image classification \cite{miyai2023locoop,li2024learning} and open vocabulary detection \cite{zhong2022regionclip,minderer2022simple}. However, there are still countless objects that remain outside the scope of the predefined vocabulary in the real world, as depicted in Fig. \ref{fig:OS&OV}(a). For instance, assuming the current label consists of the vocabulary [`car', `bird', `horse'], applying it to the two images in Fig. \ref{fig:false_missed_true} may lead to false or missed detection. Ideally, the model should distinguish between ``\textit{what is known}'' and ``\textit{what is unknown}'', as shown in Fig. \ref{fig:false_missed_true} (right). Therefore, combining prompt learning with the construction of a robust decision boundary could enhance the FOOD detector for unknown rejection.

In this paper, we develop the first prompt-based FOOD framework for unknown rejection. As no training data is available for unknown classes, we select pseudo-unknown samples from region proposals for unknown class optimization. Drawing inspiration from the gradient-based attribution method \cite{chen2024gaia} for uncertainty estimation, we propose an Attribution-Gradient based Pseudo-unknown Mining (AGPM) method to select pseudo-unknown samples. It benefits from the diverse interpretative capabilities of different texts for the same image, which is reflected in the attribution gradient differences within the large model. However, these pseudo-unknown samples often contain a mixture of known and unknown properties, which cannot fit the real unknown distribution. As shown in Fig. \ref{fig:OS&OV}(b), the region proposal could contain features of both the car within the vocabulary and the unknown tire, causing an ambiguous decision boundary between knowns and unknowns. To alleviate this problem, the proposed Conditional Evidence Decoupling (CED) method decouples known and unknown properties from pseudo-unknown samples. This approach is derived from the uncertainty mining property of Evidential Deep Learning \cite{sensoy2018evidential} while removing the evidence influence of the ground truth class, as shown in Fig. \ref{fig:OS&OV}(c). Furthermore, to reduce the impact of local abnormal features on the final decision, the proposed Abnormal Distribution Calibration (ADC) method adjusts the output probability distribution of local features to regularize the model and strengthen the unknown decision boundary. Experimental results demonstrate the superiority of our method on both known and unknown classes. We summarize our main contributions as follows:

\begin{itemize}
    \item To the best of our knowledge, this is the first work to employ prompt learning for the FOOD task. We propose an Attribution-Gradient based Pseudo-unknown Mining method by innovatively quantifying the text interpretative capabilities in the image-text joint space to mine high uncertainty region proposals.
    \item We propose Conditional Evidence Decoupling to extract distinct information from pseudo-unknown samples. To enhance robustness, we introduce Abnormality Distribution Calibration to regularize the model and shape a discriminative unknown decision boundary.
    \item Extensive experiments demonstrate the effectiveness of our FOOD method, which outperforms previous vision-only frameworks and achieves superior unknown rejection performance.
    
\end{itemize}

\section{Related Work}

\textbf{Vision Language Models} (VLMs) \cite{radford2021learning,zhong2022regionclip} leverage extensive image-text pre-training data, enabling zero-shot detection of desired classes based on given vocabulary, and they could quickly adapt to downstream tasks with prompt learning \cite{zhou2022coop,zhou2022conditional}. VLMs use predefined vocabulary to extend their understanding of the real world. However, real-world data often follows a long-tail distribution, where many rare or non-typical objects are difficult to define and cannot be effectively addressed by Open-Vocabulary Detection (OVD) \cite{zareian2021open,zhong2022regionclip,minderer2022simple,liu2023grounding}. We tackle this challenge by FOOD, which ensures accurate recognition of known classes in vocabulary while preventing false detections. It is valuable for advancing the robustness and versatility of vision-language models in handling complex and diverse real-world scenarios.

\begin{figure*}[htbp]
\centering
\includegraphics[width=0.9\linewidth]{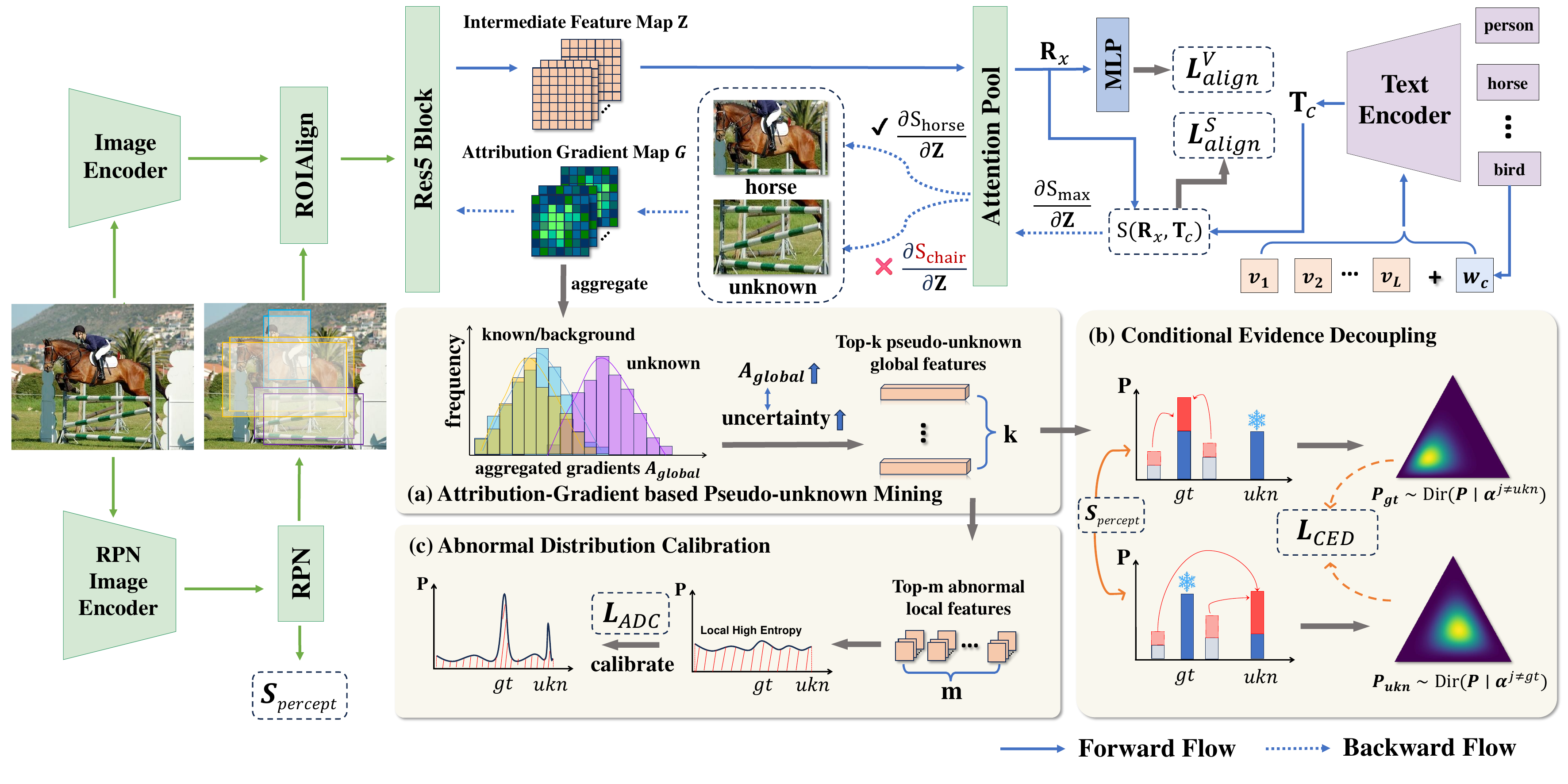}
\caption{The overview architecture of our method, which is a two-stage detector with (a) Attribution-Gradient based Pseudo-unknown Mining, (b) Conditional Evidence Decoupling For Unknown Optimization, (c) Abnormal Distribution Calibration For Robust Decision Boundary.}
\vspace{-1em}
\label{fig:pipeline}
\end{figure*}

\textbf{Pseudo-unknown Sample Mining.}
 There are no training samples for unknown classes, thus, the goal of pseudo-unknown sample mining is to select highly uncertain samples from foreground and background proposals for subsequent optimization of unknown classes. \cite{han2022expanding} used a maximum entropy for pseudo-unknown sample mining in open-set object detection. \cite{su2024toward} employed maximum conditional energy in few-shot open-set object detection, and in FOODv2 \cite{su2023hsic}, they selected proposals with high evidence uncertainty as pseudo-unknown samples. While these methods all operated within the visual feature space, we explore pseudo-unknown sample mining with additional textual information.

\textbf{Few-Shot Open-Set Object Detection.} Compared with Few-Shot Open-Set Recognition (FSOSR) \cite{liu2020few,jeong2021few,wang2023glocal,boudiaf2023open,nag2023reconstruction}, the task of few-shot open-set object detection (FOOD) required fine-grained, region-level representations and cannot overlook the impact of background region proposals for the unknown rejection. \cite{su2024toward} initially established a benchmark for the FOOD task. To enhance generalization for unknown classes, \cite{su2023hsic} proposed a Hilbert-Schmidt Independence Criterion (HSIC) based moving weight averaging technique to regularize the updating of model parameters. In this paper, we are dedicated to decoupling known and unknown information in pseudo-unknown samples with evidential deep learning to establish a robust decision boundary between known and unknown classes.

\section{Method}
Our method is a prompt-based few-shot open-set object detection framework including Attribution-Gradient based Pseudo-unknown Mining for uncertainty quantification, Conditional Evidence Decoupling for unknown optimization, and Abnormal Distribution Calibration for robust unknown decision boundary. An overview of our method is shown in Fig. \ref{fig:pipeline}. RegionCLIP is adopted \cite{zhong2022regionclip} as the base framework, composed of two image encoders, a separately trained region proposal network (RPN), and a text encoder. To alleviate the issue of traditional RPNs falsely being class-agnostic \cite{wang2023random,sarkar2024open}, we train the RPN with a parallel branch to compute the centerness score \cite{tian2020fcos}, which provides robust localization ability from object position and shape.

\subsection{Preliminary}
We formalize the FOOD task based on previous research \cite{su2024toward} as training a detector with a class-imbalanced training dataset to accurately classify the \(K\) known classes \(C_{Known}=C_{Base} \cup C_{Novel}\), reject all unknown classes \(C_{Ukn}\), and distinguish between foreground and background according to \(C_{B G}\).

Conventionally, the prompt learning method CoOp \cite{zhou2022coop} aligns the region feature \(\mathbf{R}\) with class-specific prompts \(\mathbf{T}\), where context words of prompt templates (e.g., ``a photo of a'') are replaced with continuously learnable parameters, denoted as \(\mathbf{t}_{c}=\left\{\boldsymbol{v}_{1}, \boldsymbol{v}_{2}, \ldots, \boldsymbol{v}_{L}, \boldsymbol{w}_{c}\right\}\). The semantic alignment loss is defined as:


\begin{equation}  
\resizebox{.91\linewidth}{!}{$\boldsymbol{L}_{align}^{S}=-\frac{1}{N}\sum_{i=1}^{N}\sum_{j=1}^{K+2} {y}_{i j} \log \frac{\exp \left(\textup{S}\left(\mathbf{R}_{i}, \mathbf{T}_{j}\right) / \tau\right)}{\sum_{c=1}^{K} \exp \left(\left(\textup{S}\left(\mathbf{R}_{i}, \mathbf{T}_{c}\right)\right) / \tau\right)} $},
\label{Eq:1}
\end{equation}
where \(\textup{S}\left(\cdot, \cdot\right)\) represents the cosine similarity and \(\tau\) denotes the temperature parameter, \({y}_{ij}\) is an indicator (0 or 1) of sample \(i\) belonging to category \(j\) in the ground truth label.

Following \cite{han2022expanding}, we implement enqueue/dequeue operations based on the memory bank and regularize the model with the following visual alignment loss to align the region feature with the same class:
\begin{equation}  
{\boldsymbol{L}_{align}^{V}=\frac{1}{N} \sum_{i=1}^{N} \boldsymbol{L}_{align }^{V}\left(\mathbf{z}_{i}\right) },
\label{Eq:2}
\end{equation}

\begin{equation}
\resizebox{.91\linewidth}{!}
{$\displaystyle \boldsymbol{L}_{align }^{V}\left(\mathbf{z}_{i}\right)=\frac{1}{\left|Q\left(\mathbf{c}_{i}\right)\right|} \sum_{\mathbf{z}_{j} \in Q\left(\mathbf{c}_{i}\right)} \log \frac{\exp \left(\mathbf{z}_{i} \cdot \mathbf{z}_{j} / \varepsilon\right)}{\sum_{\mathbf{z}_{k} \in Q \backslash {Q}_{\mathbf{c}_{i}} } \exp \left(\mathbf{z}_{i} \cdot \mathbf{z}_{k} / \varepsilon\right)} $},
\label{Eq:3}
\end{equation}

where \(\mathbf{z}\) is 128-dimensional latent embeddings mapped from \(\mathbf{R}\), \(\mathbf{c}_{i}\) is the class label for the \(i\)-the proposal, \(\varepsilon\) is a hyperparameter, and \(Q\left(\mathbf{c}_{i}\right)\) represents the embedding queue for class \(\mathbf{c}_{i}\).

\subsection{Attribution-Gradient based Pseudo-unknown Mining}
Due to the lack of training data for unknown classes, it is difficult to establish a clear unknown decision boundary. To tackle the above issue, we select a subset of known proposals as pseudo-unknown samples, which may exhibit features of unknown classes. Inspired by the gradient-based attribution method, which was first introduced in the sensitivity analysis (SA) \cite{simonyan2013deep}, it evaluates the sensitivity of a particular input feature on the final prediction output for visual interpretability \cite{selvaraju2017grad}. We propose a novel Attribution-Gradient based Pseudo-unknown Mining (AGPM) method to mine high-uncertainty pseudo-unknown samples, which are then employed to construct an unknown decision boundary. Specifically, we take the intermediate feature layer \(\mathbf{Z}\) (in Fig. \ref{fig:pipeline}) as the target layer. For a given proposal feature \(\mathbf{R}_{x}\), we obtain the attribution gradient at \(\text{Z}_{ij}^{k}\) corresponding to the maximum text-image matching score:
\begin{equation}  
{{G}_{ij}^{k}=\frac{\partial \max _{c=1 \ldots K} \textup{S}\left(\mathbf{R}_{x}, \mathbf{T}_{c}\right)}{\partial \text{Z}_{ij}^{k}}},
\label{Eq:4}
\end{equation}
where \(i\), \(j\), and \(k\) represent the indices of height, width, and channel, respectively. We can obtain the attribution gradient map \(\boldsymbol{G}\) corresponding to different proposals. Consequently, we perform global aggregation of attribution gradients as follows: 
\begin{equation} 
{A_{global}=\frac{1}{C} \sum_{k}^{C}\left(\sum_{i}^{H} \sum_{j}^{W} \gamma_{i j}^{k}\right) \cdot\left(\sum_{i}^{H} \sum_{j}^{W}\left|G_{i j}^{k}\right|\right)},
\label{Eq:5}
\end{equation}
where \(\gamma_{i j}^{k}\) is an indicator function such that \(\gamma_{i j}^{k}=1\) if \(G_{i j}^{k} \ne 0\) and \(\gamma_{i j}^{k}=0\) if \(G_{i j}^{k} = 0\), and \(|\cdot|\) denotes the absolute function, resulting in a scalar aggregated outcome. This result could serve as a metric for quantifying uncertainty and assessing the differences between known and unknown classes.

\begin{figure}[!t]
     \centering
     
     \subfloat[voc07\&12trainval]{\includegraphics[width=0.15\textwidth]{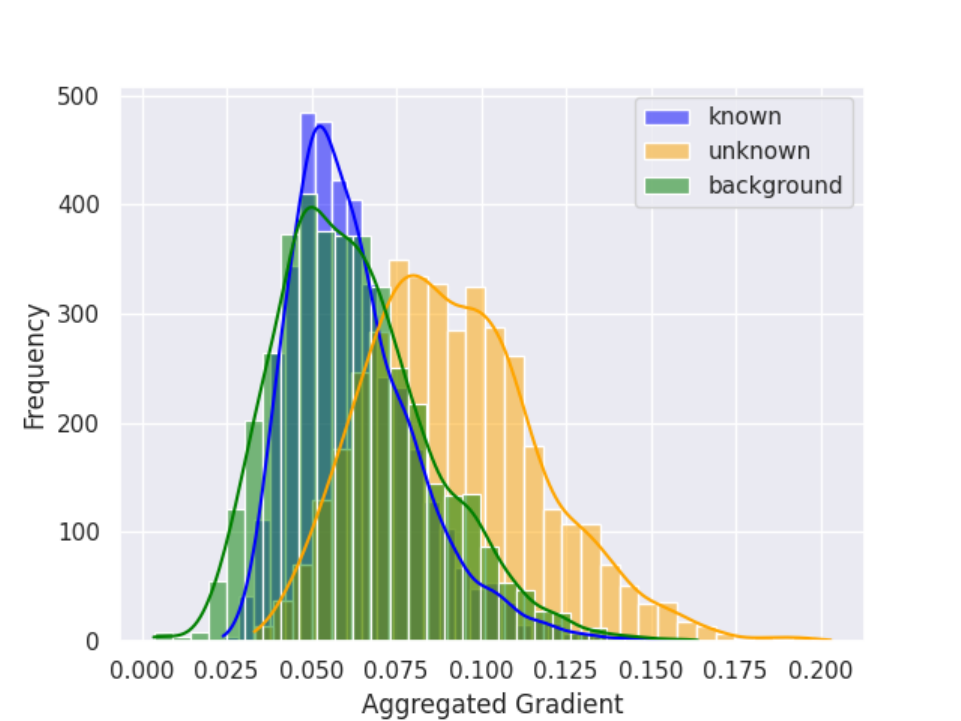}
\label{fig:y equals x}}
     \hfill
     \subfloat[voc07test]{\includegraphics[width=0.15\textwidth]{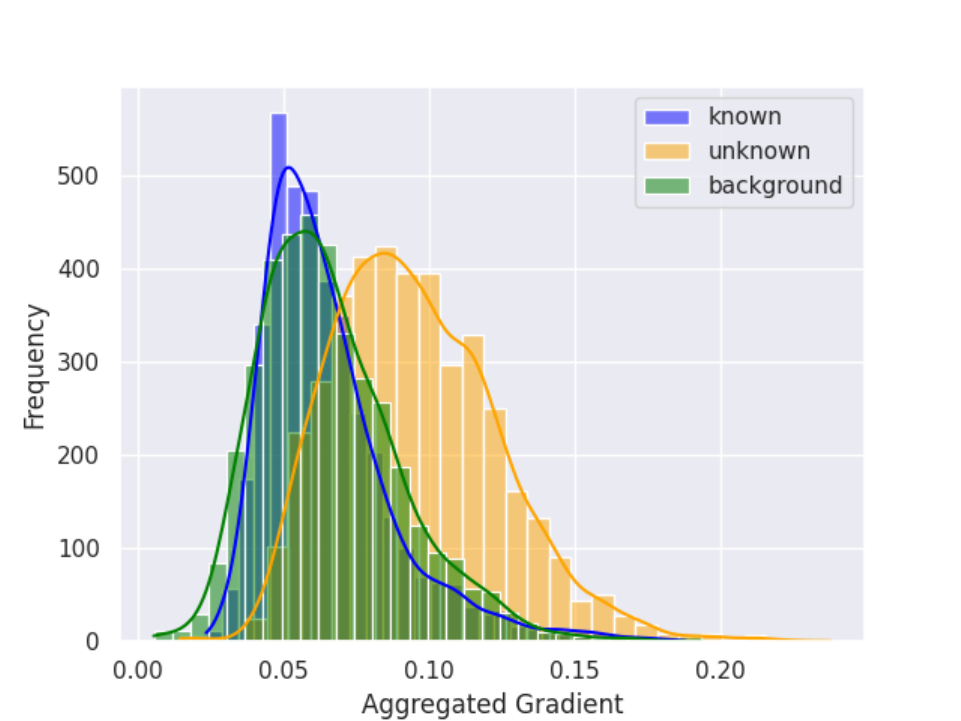}
         \label{fig:three sin x}}
     \hfill
     \subfloat[coco2017val]{\includegraphics[width=0.15\textwidth]{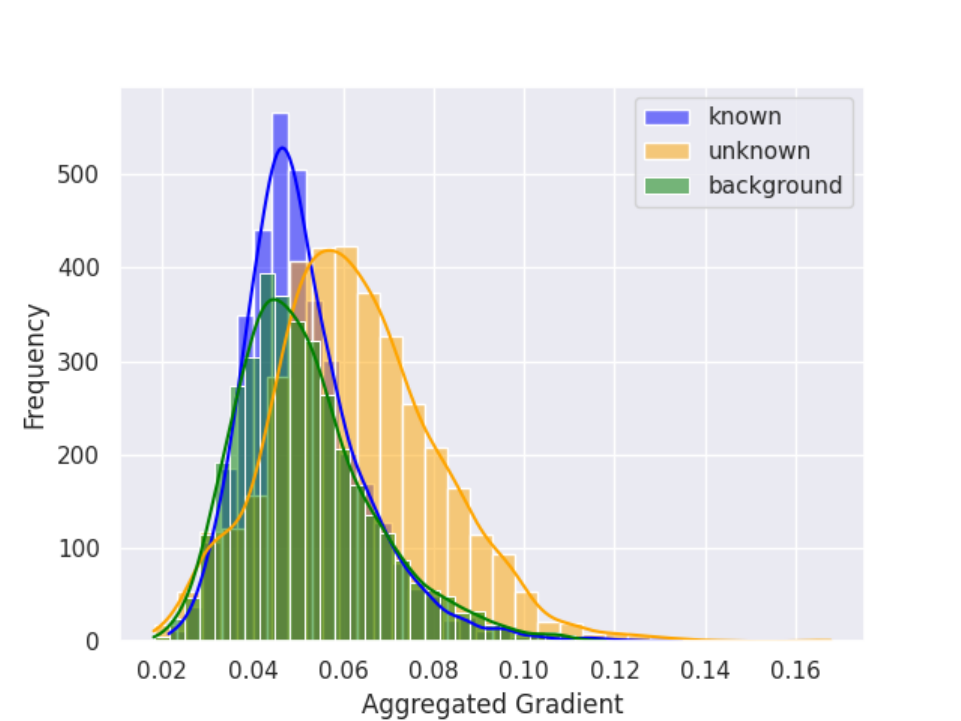}
         \label{fig:five over x}}

    \caption{Distribution of global aggregation attribution gradients across known, background, and unknown classes. Proposals are sampled from 500 randomly selected images in the VOC10-5-5 base training/testing sets and the VOC-COCO testing set, excluding the VOC-COCO training set as it contains only base class labels.}
    \label{fig:global aggre}
\end{figure}

 We then analyze the distributions of \(A_{global}\) for known, background, and unknown classes with all labels available, identifying distinct distribution patterns, as shown in Fig. \ref{fig:global aggre}. Under the premise of having only known class labels, a higher \(A_{global}\) aligns more closely with the distribution characteristic of unknown classes. Therefore, we select the proposals corresponding to the top-\(k\) highest \(A_{global}\) from the foreground and background proposals as pseudo-unknown samples with sampling ratio \(S_{fg:bg}\).

\subsection{Conditional Evidence Decoupling For Unknown Optimization}
For FOOD, the unknown objects are easily misclassified into known ones with a high confidence score, which could be attributed to its coupling of known and unknown information. To decouple and learn distinct information from the pseudo-unknown samples, we reserve a placeholder beyond the vocabulary for unknown classes and model the relationship between known and unknown classes based on conditional evidence. Specifically, we employ Evidential Deep Learning (EDL) \cite{sensoy2018evidential} based on the evidence framework of Dempster-Shafer Theory (DST) \cite{sentz2002combination} and the subjective logic (SL) \cite{josang2016subjective} to estimate uncertainty. By assuming that output probabilities \(\boldsymbol{P}\) follow a Dirichlet distribution, denoted as \(\boldsymbol{P} \sim \operatorname{Dir}\left(\boldsymbol{P} \mid \boldsymbol{\alpha}\right)\), EDL constructs distribution of distributions for uncertainty modeling. Drawing on the DST and SL theory, for a classifier with \(K+2\) classes, we denote \(\exp (l_{i}^{j})\) as the evidence output for the \(j\)-th class from the \(i\)-th proposal, where \(l_{i}^{j}=\textup{S}\left(\mathbf{R}_{i}, \mathbf{T}_{j}\right) / \tau\). Consequently, this allows deriving the parameters for the Dirichlet distribution:
\begin{equation}  
{\alpha_{i}^{j}=\exp(l_{i}^{j})+1}.
\label{Eq:7}
\end{equation}  

To extract distinct knowledge from identical features, we optimize evidence for known and unknown classes separately. Furthermore, we eliminate the evidence of the ground-truth class while optimizing for the unknown class, and conversely for known classes. In this case, we can alleviate the performance degradation caused by the contradictory evidence of decoupled classes. We formalize this as a conditional EDL loss in the following form:
\begin{equation}  
{\boldsymbol{L}_{i}^{ukn}=\psi\left(\sum_{j = 1, j \neq g t}^{K+2} \alpha_{i}^{j}\right)-\psi\left(\alpha_{i}^{ukn}\right)},
\label{Eq:8}
\end{equation}
\begin{equation}  
{\boldsymbol{L}_{i}^{g t}=\psi\left(\sum_{j =1, j \neq ukn}^{K+2} \alpha_{i}^{j}\right)-\psi\left(\alpha_{i}^{g t}\right)},
\label{Eq:9}
\end{equation}
where \(\psi(\cdot)\) represents the digamma function, \(\boldsymbol{L}^{ukn}\) and \(\boldsymbol{L}^{gt}\) optimize the evidence for the known and unknown classes, respectively. Subsequently, we use the object perception score as a weight factor to balance the optimization between known and unknown classes, which is derived from the geometric mean of the original objectness score \(S_{obj}\) and the centerness score \(S_{center}\) from RPN:
\begin{equation}  
{S_{percept}=\sqrt{S_{obj}\cdot S_{center}} }.
\label{Eq:6}
\end{equation}

Intuitively, a higher \(S_{percept}\) of foreground proposals indicates more known information, thereby increasing the weight for optimizing known classes, conversely for background proposals. Consequently, we derive the following foreground and background conditional evidence decoupling losses:
\begin{equation}  
{\boldsymbol{L}_{CED}^{f g}=\frac{1}{N} \sum_{i=1}^{N}\left(1-{S_{percept}}_{i}\right) \cdot L_{i}^{ukn}+{S_{percept}}_{i} \cdot L_{i}^{g t}},
\label{Eq:10}
\end{equation}
\begin{equation}  
{\boldsymbol{L}_{CED}^{b g}=\frac{1}{N} \sum_{i=1}^{N} {S_{percept}}_{i} \cdot L_{i}^{ukn}+\left(1- {S_{percept}}_{i}\right) \cdot L_{i}^{g t}}.
\label{Eq:11}
\end{equation}

To this end, the final loss expression is as follows:
\begin{equation}  
{\boldsymbol{L}_{CED}=\boldsymbol{L}_{CED}^{f g}+\boldsymbol{L}_{CED}^{b g}}.
\label{Eq:12}
\end{equation}

\begin{figure}[!t]
     \centering
     \subfloat[known]{\includegraphics[width=0.16\textwidth]{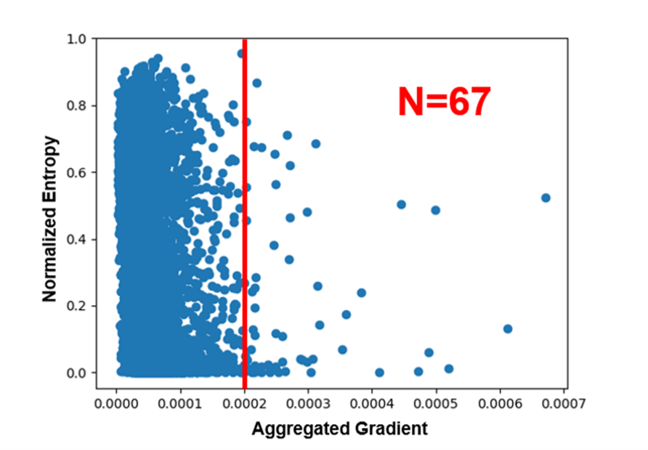}
         \label{fig:known}}
     \subfloat[background]{\includegraphics[width=0.16\textwidth]{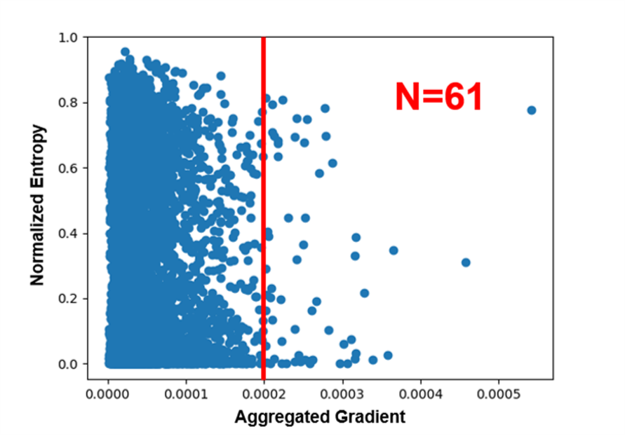}
         \label{fig:background}}
     \subfloat[unknown]{\includegraphics[width=0.16\textwidth]{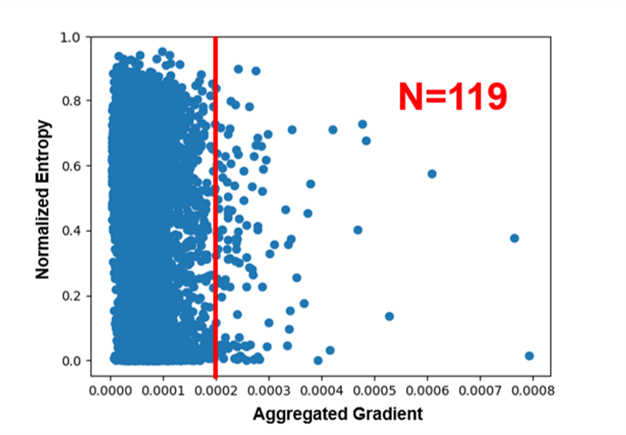}
         \label{fig:unknown}}
    \caption{Scatter plots of known, background, and unknown classes on the VOC07+12trainval dataset. Each point represents a local feature \(\mathbf{Z}_{xy}\) from intermediate output \(\mathbf{Z}\). Proposals of unknown classes show twice as many gradient outliers (threshold \(>\) 0.0002) as known and background classes, with 100 selected proposals per plot.}
    \label{fig:local_aggre}
\end{figure}

\subsection{Abnormal Distribution Calibration For Robust Decision Boundary}
 By employing conditional evidence decoupling, the detector can distinguish known and unknown classes using global features. However, certain local anomalous features \(\mathbf{Z}_{xy}\) still pose a disruption to the decision-making process. Therefore, we delve into the reasons for the differences in global attribution gradient distributions by aggregating local attribution gradients. We observe that, compared to known and background classes, unknown classes exhibit a greater number of outliers in locally aggregated attribution gradients, as shown in Fig. \ref{fig:local_aggre}. For the attribution gradient map \(G\), we perform aggregation along the channel dimension, resulting in local aggregation as follows:
\begin{equation}  
{A_{local}=\frac{1}{C} \sum_{k}^{C}\left|G_{xy}^{k}\right|},
\label{Eq:13}
\end{equation}
for each local position \((x, y)\), \(A_{local}\) is a scalar, \(C\) is the total number of channels. We posit that outlier gradients reflect uncertain local features that hinder global discrimination between known and unknown classes. To address this, we recalibrate the probability distribution by reducing non-ground-truth logits to mitigate overconfidence. Specifically, we project the pseudo-unknown local features \(\mathbf{Z}_{xy}\) into the image-text joint space. The match scores between local and textual features are then computed to obtain the local output logits \(l{'}\). These logits are then adjusted using the following abnormality distribution calibration loss to recalibrate the local output distribution:
\begin{equation}  
\begin{split}
\label{Eq:14}
    \boldsymbol{L}_{ADC}=-\frac{1}{M} \sum_{i}^{M} ( \sum_{j =1 , j \neq g t}^{K} \log \frac{\exp(-l{'}_{i}^{j})}{1+\exp(-l{'}_{i}^{j})}\\
+H_{norm}(\boldsymbol{p}{'})\cdot \log \frac{1}{1+\exp(-l{'}_{i}^{ukn})} ),
\end{split}
\end{equation}
where \(H_{norm}(\boldsymbol{p})=- {\textstyle \sum_{c}p{'}_{c}\log {p{'}_{c}}}/\log (K)\) represents the normalized entropy, serving as a weighting factor to constrain the learning of the unknown class. For each pseudo-unknown sample, we select the top-\(m\) highest \(A_{local}\) local features to recalibrate the output probability distribution,

\subsection{Overall Optimization}
We adopt a two-stage fine-tuning strategy \cite{wang2020frustratingly} to train the few-shot open-set detector, for the base training stage:
\begin{equation}  
{\boldsymbol{L}_{base}=\boldsymbol{L}_{reg}+\boldsymbol{L}_{align}^{S}+ \gamma_{t} \boldsymbol{L}_{align}^{V}},
\label{Eq:15}
\end{equation}
and for the few-shot fine-tuning stage:
\begin{equation}
\resizebox{.91\linewidth}{!}
{$\boldsymbol{L}_{novel}=\boldsymbol{L}_{reg}+\boldsymbol{L}_{align}^{S}+ \gamma_{t} \boldsymbol{L}_{align}^{V}+\lambda_{t}(\boldsymbol{L}_{CED}+\beta\boldsymbol{L}_{ADC})$},
\label{Eq:16}
\end{equation}
where \(\boldsymbol{L}_{reg}\) is smooth L1 loss for box regression, \(\gamma_{t}\) is a stepwise decreasing weight strategy similar to \cite{han2022expanding}, \(\beta\) is a hyperparameter and \(\lambda_{t}=\exp(\log (\lambda )\cdot (1-t/T)) \in [\lambda, 1]\) denotes the weight that changes exponentially with the current iteration (\(t\)) and the total iteration (\(T\)), of which intention is to first learn well-defined semantic clusters and then gradually establish decision boundary between known and unknown classes.

\begin{table*}
\vspace{-0.5em}
  \centering
 \scriptsize

  \begin{tabular}{lcccccc}
    \toprule
    \multirow{2}{*}{} & \multicolumn{3}{c}{1-shot} & \multicolumn{3}{c}{3-shot} \\
    \cmidrule(l){2-4} \cmidrule(l){5-7}
     {Method}& {\(mAP_{K} \)}  / {\(mAP_{N}\uparrow \)} & {\(R_{U}\)} / {\(AR_{U}\uparrow \)} & {\(WI \)} / {\(AOSE\downarrow \)}
     & {\(mAP_{K} \)}  / {\(mAP_{N}\uparrow \)} & {\(R_{U}\)} / {\(AR_{U}\uparrow \)} & {\(WI \)} / {\(AOSE\downarrow \)}
     \\
    \midrule
    ORE \cite{joseph2021towards}    
    & 43.25  / 8.62& 18.25 / {--}  & 9.54  / 930.30 
                & 45.88  / 14.52& 22.23 / {--}  & 9.88  / 1058.70
    \\
    PROSER \cite{zhou2021learning} 
    & 41.64 / 8.49& 30.95 / 15.41 & 11.15 / 994.60 
                & 43.30  / 15.16& 32.30 / 16.17  & 10.45  / 1021.70
    \\
    OPENDET \cite{han2022expanding} 
    & 43.45 / 8.27& 33.64 / 17.28 & 10.47 / 867.30 
                & 46.47  / 14.09& 30.62 / 15.89  & 9.27  / 954.50
    \\
    FOOD \cite{su2024toward}        
    & 43.97 / 8.95& 43.72 / 23.51 & 6.96  / 598.60 
                & 48.48  / 16.83& 44.52 / 23.58  & 7.83  / 859.00
    \\
    FOODv2 \cite{su2023hsic}        
    & 45.12 / 11.56 & 60.03 / 31.19 & {--}  / {--}  
                & 48.90  / 18.96& 61.21 / 32.02  & {--}  / {--}
    \\
    \midrule
    OPENDET\textbf{(+Ours)}        
    & 50.28 / 18.40 & \underline{78.56} / \underline{36.76}  & 5.89 / \underline{781.60}  
                & \underline{55.61}  / \underline{33.03}& 78.87 / \underline{38.43}  & \underline{4.75}  / \underline{547.20}
    \\
    FOODv2\textbf{(+Ours)}         
    & \textbf{53.71} / \textbf{22.62} & 77.28 / 34.70 & \underline{5.65}  / 1042.20  
                &  \textbf{56.53} / \textbf{35.65}& \underline{80.19} / 36.80  & 5.52  / 949.80
    \\
    \textbf{Ours}        & \underline{51.94} / \underline{21.43} & \textbf{79.88} / \textbf{38.12} & \textbf{4.12}  / \textbf{459.60} 
                & 53.09 / 31.70 & \textbf{80.55} / \textbf{39.53} & \textbf{3.72}  / \textbf{451.20} 
    \\
    \midrule[1pt]

    \multirow{2}{*}{} & \multicolumn{3}{c}{5-shot} & \multicolumn{3}{c}{10-shot} \\
    \cmidrule(l){2-4} \cmidrule(l){5-7}
     {Method}& {\(mAP_{K} \)}  / {\(mAP_{N}\uparrow \)} & {\(R_{U}\)} / {\(AR_{U}\uparrow \)} & {\(WI \)} / {\(AOSE\downarrow \)}
     & {\(mAP_{K} \)}  / {\(mAP_{N}\uparrow \)} & {\(R_{U}\)} / {\(AR_{U}\uparrow \)} & {\(WI \)} / {\(AOSE\downarrow \)}
     \\
    \midrule
    ORE \cite{joseph2021towards}     
    & 46.29  / 18.49& 23.01 / {--}  & 10.16  /1019.70 
                & 48.17  / 25.40& 23.48 / {--}  & 9.65  / 1063.70
    \\
    PROSER \cite{zhou2021learning}  
    & 45.12 / 20.08& 32.68 / 16.48 & 10.65 / 1009.80 
                & 48.35  / 25.13& 32.61 / 17.01  & 10.29  / 956.70
    \\
    OPENDET \cite{han2022expanding} 
    & 47.56 / 17.90& 32.13 / 16.72 & 9.01 / 1031.50 
                & 50.95 / 25.14& 36.30 / 18.89  & 8.50 / 1021.40
    \\
    FOOD \cite{su2024toward}        
    & 50.18 / 23.10& 45.65 / 23.61 & 7.59  / 908.00 
                & 53.23  / 28.60& 45.84 / 23.86  & 6.99  / 900.20
    \\
    FOODv2 \cite{su2023hsic}        
    & 52.55 / 27.31 & 62.02 / 32.79 & {--}  / {--}  
                & 57.24  / 32.63&  62.14 / 32.80  &  {--} / {--} 
    \\
    \midrule
    OPENDET\textbf{(+Ours)}        
    & \textbf{56.01}/ 36.57 & 79.70 / \underline{39.42} & \underline{4.53}  / \underline{519.40}  
                &  \underline{58.70} / 42.69& 74.60 / 37.16  &  4.90 / \textbf{530.60}
    \\
    FOODv2\textbf{(+Ours)}         
    & \underline{55.13} / \textbf{38.28} & \underline{80.62} / 37.05 & 4.98  / 1185.60  
                &  \textbf{60.84} / \textbf{45.56}& \textbf{79.45} / \underline{37.17}  & \underline{4.12}  / 953.30
    \\
    \textbf{Ours}        & 54.35 / \underline{36.67} & \textbf{81.37} / \textbf{40.32} & \textbf{3.78}  / \textbf{512.20} 
                & 58.55 / \underline{43.52} & \underline{79.39} / \textbf{39.79} & \textbf{3.43}  / \underline{546.30} 
    \\
    \bottomrule

  \end{tabular}
        \caption{Few-Shot Open-Set Object Detection results on VOC10-5-5. `\textbf{(+Ours)}' indicates the implementation with our prompt-based FOOD framework while `\textbf{Ours}' denotes our framework with all of our methods. \textbf{Bold} indicates the best, \underline{underlined} indicates the second best.}

  \label{tab:t1}
\vspace{-1em}
\end{table*}

\begin{table*}[!h]

  \centering
    \scriptsize

  \begin{tabular}{lcccccc}
    \toprule
    \multirow{2}{*}{} & \multicolumn{3}{c}{1-shot} & \multicolumn{3}{c}{5-shot} \\
    \cmidrule(l){2-4} \cmidrule(l){5-7}
     {Method}& {\(mAP_{K} \)}  / {\(mAP_{N}\uparrow \)} & {\(R_{U}\)} / {\(AR_{U}\uparrow \)} & {\(WI \)} / {\(AOSE\downarrow \)}
     & {\(mAP_{K} \)}  / {\(mAP_{N}\uparrow \)} & {\(R_{U}\)} / {\(AR_{U}\uparrow \)} & {\(WI \)} / {\(AOSE\downarrow \)}
     \\
    \midrule
    ORE \cite{joseph2021towards} 
    & 14.14  / 2.18& 4.59 / {--}  & 12.08  / 1087.00 
                & 16.21  / 6.29& 4.99 / {--}  & 12.30  / 1344.00
    \\
    PROSER \cite{zhou2021learning} 
    & 13.58 / 2.32& 7.53 / 3.07 & 11.68 / 925.30 
                & 15.67  / 6.40& 9.59 / 4.08  & 12.56  / 1165.90
    \\
    OPENDET \cite{han2022expanding}
    & 16.01 / 2.29& 7.24 / 3.14 & 9.82 /  690.90
                & 17.16  / 6.56& 11.49 / 5.21  & 9.55  / 1176.90
    \\
    FOOD \cite{su2024toward}
    & 15.83 / 2.26& 15.76 / 7.20 & 6.78  / \textbf{485.00} 
                & 18.08  / 6.69& 20.02 / 9.45  & 7.37  / 859.00
    \\
    FOODv2 \cite{su2023hsic}
    & 18.54 / 4.33 & 30.87 / 14.13 & {--}  / {--}  
                & 19.88  / 11.95& 32.53 / 15.74  & {--}  / {--}
    \\
    \midrule
    OPENDET\textbf{(+Ours)}        
    & 18.42 / 4.42 & \underline{36.70} / \underline{16.17} & 5.42  / 796.80  
                & 20.42 / 12.23 & \underline{39.10} / \underline{17.89} & 4.83  / \textbf{742.40}
    \\
    FOODv2\textbf{(+Ours)}        
    & \textbf{20.44} / \textbf{5.69} & 36.25 / 15.74 & \underline{5.14}  / 945.40  
                & \underline{21.12}  / \underline{12.47}& 39.05 / 16.72  & \underline{4.70}  / 835.90
    \\
    \textbf{Ours}        & \underline{19.49} / \underline{5.41} & \textbf{38.53} / \textbf{16.68} & \textbf{4.51}  / \underline{638.70} 
                & \textbf{21.46} / \textbf{13.24} & \textbf{40.52} / \textbf{17.91} & \textbf{2.99}  / \underline{808.90} 
    \\
    \midrule[1pt]

    \multirow{2}{*}{} & \multicolumn{3}{c}{10-shot} & \multicolumn{3}{c}{30-shot} \\
    \cmidrule(l){2-4} \cmidrule(l){5-7}
     {Method}& {\(mAP_{K} \)}  / {\(mAP_{N}\uparrow \)} & {\(R_{U}\)} / {\(AR_{U}\uparrow \)} & {\(WI \)} / {\(AOSE\downarrow \)}
     & {\(mAP_{K} \)}  / {\(mAP_{N}\uparrow \)} & {\(R_{U}\)} / {\(AR_{U}\uparrow \)} & {\(WI \)} / {\(AOSE\downarrow \)}
     \\
    \midrule
    ORE \cite{joseph2021towards}    
    & 17.98  / 8.75& 5.13 / {--}  & 11.65  / 1463.20 
                & 23.07  / 15.17& 5.51 / {--}  & 11.22  / 1867.00
    \\
    PROSER \cite{zhou2021learning}  
    & 17.00 / 8.75& 10.06 / 4.89 & 12.47 / 1160.00 
                & 21.44  / 14.30& 12.06 / 5.98  & 12.00  / 1561.60
    \\
    OPENDET \cite{han2022expanding} 
    & 18.53 / 8.70& 13.89 / 6.32 & 9.83 / 1400.60 
                & 22.93 / 14.02& 18.07 / 8.76  & 9.02 / 1818.00
    \\
    FOOD \cite{su2024toward}
    & 20.17 / 9.48& 21.48 / 9.56 & 7.59  / 1099.30 
                & 23.90  / 14.17& 23.17 / 11.45  & 8.13  / 1480.00
    \\
    FOODv2 \cite{su2023hsic} 
    & 22.64 / 13.82 & 32.78 / 16.52 & {--}  / {--}  
                & 23.71  / 17.67&  35.74 / 17.26 &  {--} / {--} 
    \\
    \midrule
    OPENDET\textbf{(+Ours)}        
    & 22.74 / 15.34 & \underline{38.12} / \textbf{17.72} & 5.12  / \underline{934.30}  
                &  25.34 / \underline{21.56}& 38.78 / 16.68  & 4.97  / 1463.6
    \\
    FOODv2\textbf{(+Ours)}      
    & \textbf{24.42} / \textbf{16.83} & 37.33 / 16.04 & \underline{4.38}  / 1046.60 
                & \textbf{26.70}  / \textbf{22.73}& \textbf{39.46} / \underline{17.24}  & \underline{4.23}  / \underline{1442.40}
    \\
    \textbf{Ours}        & \underline{23.75} / \underline{16.77} & \textbf{38.69} / \underline{17.06} & \textbf{2.58}  / \textbf{856.40} 
                & \underline{25.72} / 21.16 & \underline{39.43} / \textbf{17.52} & \textbf{2.46}  / \textbf{1339.30} 
    \\
    \bottomrule
  \end{tabular}
    \caption{Few-Shot Open-Set Object Detection results on VOC-COCO. `\textbf{(+Ours)}' indicates the implementation with our prompt-based FOOD framework while `\textbf{Ours}' denotes our framework with all of our methods. \textbf{Bold} indicates the best, \underline{underlined} indicates the second best.}

  \label{tab:t2}
\end{table*}

\section{Experiments}
\subsection{Experimental Details} \label{Experimental Detail}

\textit{1) Datasets:} Following \cite{su2024toward}, the data splits VOC10-5-5, VOC-COCO, and COCO-RoadAnomaly \cite{lis2019detecting} are used for performance evaluation. \textbf{VOC10-5-5} includes 10 base, 5 novel, and 5 unknown classes from PASCAL VOC \cite{everingham2010pascal}. \textbf{VOC-COCO} has 20 base classes from PASCAL VOC, 20 novel classes from non-overlapping MS COCO \cite{lin2014microsoft}, and 40 unknown classes. \textbf{COCO-RoadAnomaly} evaluates model generalization in open-set road scenes.    

\textit{2) Evaluation Metrics:} For the FOOD task, the mean Average Precision (\(mAP\)) of known classes (\(mAP_{K}\)) and novel classes (\(mAP_{N}\)) are adopted as known class metrics. For unknown class metrics, the recall (\(R_{U}\)) and average recall (\(AR_{U}\)) of unknown classes are reported as in \cite{su2023hsic}. Furthermore, we report Wilderness Impact (\(WI\)) under a recall level of 0.8 to measure the degree of unknown objects misclassified to known ones: \(WI=\frac{P_{K}}{P_{K\cup U}} - 1\), and Absolute Open-Set Error (\(AOSE\)) to count the number of misclassified unknown objects as in \cite{han2022expanding}.

\textit{3) Implementation Details:} We employ RegionCLIP \cite{zhong2022regionclip} as the image encoder, and ResNet-50 \cite{he2016deep} pre-trained on ImageNet as the RPN image encoder. Class-specific prompt training follows CoOp \cite{zhou2022coop} with a context length of 16, using a two-stage training strategy \cite{wang2020frustratingly}, where the base training phase trains a base detector using \(C_{base}\), and the few-shot fine-tuning phase refines the model with a small, balanced set comprising both \(C_{base}\) and \(C_{novel}\). We employ SGD with 0.9 momentum, 5e-5 weight decay, and a batch size of 1 on a GTX 1080 Ti GPU. The learning rate is 0.0002 for base training and 0.0001 for fine-tuning. Visual alignment loss settings follow the setting in \cite{han2022expanding}. Other hyperparameters \(\tau\), \(\varepsilon\), \(\lambda\), and \(\beta\) are 0.01, 0.1, 1e-4, and 1.0, respectively.

\textit{4) Baselines:} We compare vision only open-set framework including: ORE \cite{joseph2021towards},PROSER \cite{zhou2021learning},OPENDET \cite{han2022expanding},FOOD \cite{su2024toward},and FOODv2 \cite{su2023hsic}. Furthermore, we employ OPENDET and FOODv2 within our open-set detection framework, denoted by OPENDET\textbf{(+Ours)} and FOODv2\textbf{(+Ours)} to conduct a fair comparison. We employ max entropy and max evidential uncertainty as pseudo-unknown sampling methods respectively, with unknown probability loss and IoU loss as optimization strategies for unknown classes respectively.

\vspace{-0.1cm}
\subsection{Main results}
\textit{1) Experiments on VOC10-5-5:} Table \ref{tab:t1} presents the FOOD results on VOC10-5-5, where we report the results of fine-tuning on 1, 3, 5, and 10 shots, averaging ten runs per setting for a fairer comparison. Based on our framework, both OPENDET\textbf{(+Ours)} and FOODv2\textbf{(+Ours)} show significant improvements compared to their original versions, demonstrating the advantages of our open-set framework. Compared to previous state-of-the-art methods with traditional open-set framework, our approach (with \(k=3, S_{fg:bg}=1:3, m=1\)) achieves significant improvement on unknown class metrics, surpassing the second best (mean results) by \textbf{18.95\%} \(R_{U}\), \textbf{7.24\%} \(AR_{U}\), \textbf{3.58} \(WI\) and \textbf{324.13} \(AOSE\). Our method consistently achieves the state-of-the-art performance on most metrics related to unknown classes, despite a slight decrease in accuracy for known classes.

\begin{figure*}[htbp]
\centering
\includegraphics[width=1\textwidth]{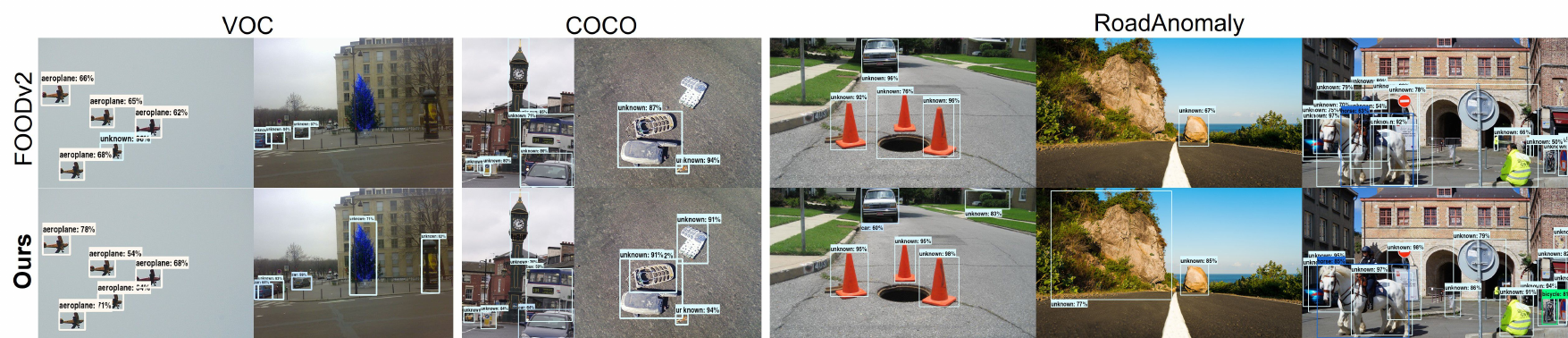}
\caption{The visualized results on VOC, COCO, and RoadAnomaly datasets under 10-shot VOC10-5-5 setting. Our method recalls more unknown objects and better distinguishes between the knowns and the unknowns.}
\label{fig:visualized}

\end{figure*}

\textit{2) Experiments on VOC-COCO:} Table \ref{tab:t2} displays the FOOD results on VOC-COCO, which is a more challenging dataset. We report results of fine-tuning on 1, 5, 10, and 30 shots, averaging ten runs per shot setting to ensure a fair comparison. Both OPENDET\textbf{(+Ours)} and FOODv2\textbf{(+Ours)} also show significant improvements compared to the original frameworks.  Compared to prior state-of-the-art methods with the traditional open-set framework, our approach (with \(k=3, S_{fg:bg}=1:1, m=1\)) shows a marked improvement, outperforming the second best (mean results) by \textbf{6.31\%} \(R_{U}\), \textbf{1.38\%} \(AR_{U}\), \textbf{4.42} \(WI\) and \textbf{70.00} \(AOSE\). The 1-shot \(AOSE\) performance does not surpass previous benchmarks, likely due to the strong learning capability of prompt-based methods with limited samples, which incurs overfitting of known classes. Our method generally achieves higher \(AR_U\) and lower \(WI\) and \(AOSE\), effectively rejecting unknown objects rather than misclassifying them as known.

	\begin{table}
        \centering
        \scriptsize
            \tabcolsep=0.05cm
        \begin{tabular}{cccc|ccccc }
        \toprule
         \(\boldsymbol{L}_{align}^{S}\) &\(\boldsymbol{L}_{align}^{V}\)&\(\boldsymbol{L}_{CED}\) & \(\boldsymbol{L}_{ADC}\) & {\(WI\downarrow\)} & {\(AOSE\downarrow\)} & {\(mAP_{K}\uparrow\)} & {\(R_{U}\uparrow\)} & {\(AR_{U}\uparrow\)} \\
        \midrule
        {\CheckmarkBold}&{}&{}&{}           & 7.06  & 2314.10 & 57.79 & 0.00 & 0.00 \\
        {\CheckmarkBold}&{\CheckmarkBold}&{}&{}      & 6.83  & 2169.10 & \textbf{58.66} & 0.00 & 0.00 \\
        {\CheckmarkBold}&{}&{\CheckmarkBold}&{}           & 5.17 & 615.80 & 49.76 & 79.63 & 37.96 \\
        {\CheckmarkBold}&{}&{\CheckmarkBold}&{\CheckmarkBold}& 4.95  & 567.60  & 51.93 & 79.38 & 37.94 \\
        
        {\CheckmarkBold}&{\CheckmarkBold}&{\CheckmarkBold}&{} & 4.69  & 539.10  & 52.02 & \textbf{79.93} & 38.02 \\
        {\CheckmarkBold}&{\CheckmarkBold}&{\CheckmarkBold}&{\CheckmarkBold} & \textbf{4.12}  & \textbf{459.60}  & 51.94 & 79.88 & \textbf{38.12} \\
        \bottomrule
        \end{tabular}
                \caption{The ablation study of proposed components.}

        \label{tab:t3}
        \end{table}

\begin{table}
\centering
\scriptsize
\tabcolsep=0.06cm

\begin{tabular}{l|ccccc}
\toprule
Method & \(WI\downarrow \) & \(AOSE\downarrow \) & \(mAP_{K}\uparrow\) & \(R_U\uparrow\) & \(AR_U\uparrow\) \\
\midrule  
OPENDET \cite{han2022expanding} & 10.47 & \underline{867.30} & 43.45 & 33.64 & 17.28 \\
OPENDET\textbf{(+Ours)}  & \underline{5.89} &  \textbf{781.60}  & \underline{50.28} & \underline{78.56} & \underline{36.76} \\
FOODv2 \cite{su2023hsic} & - & - & 45.12 & 60.03 & 31.19 \\
FOODv2\textbf{(+Ours)} & \textbf{5.65} & 1042.20  & \textbf{53.71} & \textbf{77.28} & \textbf{34.70} \\
\midrule
FOODv1\textbf{(+Ours)}(P) \cite{su2024toward} & 4.57 & 896.00 & 52.39 & 78.90 & 35.54 \\
FOODv2\textbf{(+Ours)}(P) & 4.87 & \underline{884.00} & \underline{52.95} & 78.79 & 34.35 \\
GAIA-Z\textbf{(+Ours)} \cite{chen2024gaia} & \underline{4.47} & 948.60 & \textbf{53.69} & \underline{79.64} & \underline{35.86} \\
\textbf{Ours(AGPM)} & \textbf{4.12} & \textbf{459.60} & 51.94 & \textbf{79.88} & \textbf{38.12} \\
\midrule
        
        FOODv2\textbf{(+Ours)}(U) & 5.34 & 706.10 & \textbf{52.58} & 79.32 & 37.35 \\
        Non-Decoupled   & 4.22  & \textbf{424.20} & 50.67          & 79.38  & 37.87 \\
        Non-Conditional    & \textbf{3.73}  & 538.40  & 48.98          & \textbf{82.12} & \textbf{38.45} \\
        \textbf{Original \(\boldsymbol{L}_{CED}\)}  & \underline{4.12}  & \underline{459.60} & \underline{51.94} & \underline{79.88}  & \underline{38.12} \\
\bottomrule
\end{tabular}
\caption{The main contributions analysis.}
\label{detail_analy}
\vspace{-1em}
\end{table}

\subsection{Ablation Studies}

 We ablate the proposed losses with the 1-shot VOC 10-5-5 setting in Tab. \ref{tab:t3}. By employing attribution gradients to filter pseudo-unknown samples, the proposed \(\boldsymbol{L}_{CED}\) establishes a discriminative decision boundary between known and unknown classes through decoupled evidential learning. The regularization with \(\boldsymbol{L}_{ADC}\) yields improved \(WI\) and \(AOSE\) without adversely affecting \(mAP_{K}\), indicating its facilitation for the decision boundary construction.

We conduct a more detailed ablation analysis in Tab. \ref{detail_analy}, `\textbf{(+Ours)}' denotes our open-set detection framework, while `(P)' and `(U)' represent the \textbf{P}seudo-unknown sample mining method and the \textbf{U}nknown class optimization method from the corresponding paper, respectively.
The top section demonstrates the advantages of our framework. The middle section reports the experimental results of different pseudo-unknown sample mining methods. Our AGPM shows better overall performance. The bottom section presents the results of various optimization methods for unknown classes. When we apply only Eq. \ref{Eq:8} to optimize unknown classes (Non-Decoupled), it results in a degradation of accuracy for known classes. Subsequently, we remove the conditions \(j \ne gt\) in Eq. \ref{Eq:8} and \(j\ne ukn\) in Eq. \ref{Eq:9} (Non-Conditional), which significantly reduces \(mAP_K\). By retaining the conditions, we achieve a better trade-off between known and unknown metrics. For details on the parameter ablation study, please refer to the appendix.

\subsection{Visualized results}
We conduct visual comparisons between FOODv2 \cite{su2023hsic} and our proposed method in Fig. \ref{fig:visualized} under 10-shot VOC10-5-5 experimental setup. It reveals that our method successfully recalls more unknown objects across three open-set datasets and makes more accurate distinctions between known and unknown objects. For the VOC dataset, the two images on the left illustrate that FOODv2 mistakenly classifies an airplane as an unknown object, while our method correctly identifies it. The right demonstrates that our approach successfully detects the car within the vocabulary and rejects the unknown classes (tree and billboard). This suggests that our approach enables enhanced perception of object presence and facilitates the unknown rejection.

\section{Conclusion}
In this paper, we propose a novel approach for addressing the challenging few-shot open-set detection problem. By introducing prompt learning to the FOOD task and incorporating an unknown class placeholder, our framework captures information beyond the known vocabulary. To address the unavailable training data of unknown classes, an attribution-gradient based method is proposed to mine high-uncertainty samples as pseudo-unknowns. Additionally, we propose a conditional evidence decoupling loss and a local abnormal distribution calibration loss to improve unknown class optimization and establish a robust decision boundary for unknown rejection. Extensive experiments demonstrate that our method achieves state-of-the-art performance, surpassing existing approaches.

\section*{Acknowledgments}
This paper is supported by the National Key Research and Development Program of China (Grant No. 2024YFB3310904), the Science and Technology Project of Hainan Provincial Department of Transportation (Grant No. HNJTT-KXC-2024-3-22-02), the National Natural Science Foundation of China (Grant No. 62206184, 62272018, 62372448), the Basal Research Fund of Central Public Research Institute of China (Grant No.246Z4306G) and Zhongguancun Laboratory.

\bibliographystyle{named}
\bibliography{ijcai25}

\newpage

\appendix

\section{Appendix}

\subsection{Analysis}

\begin{figure}
		\centering
		\includegraphics[width=0.45\textwidth]{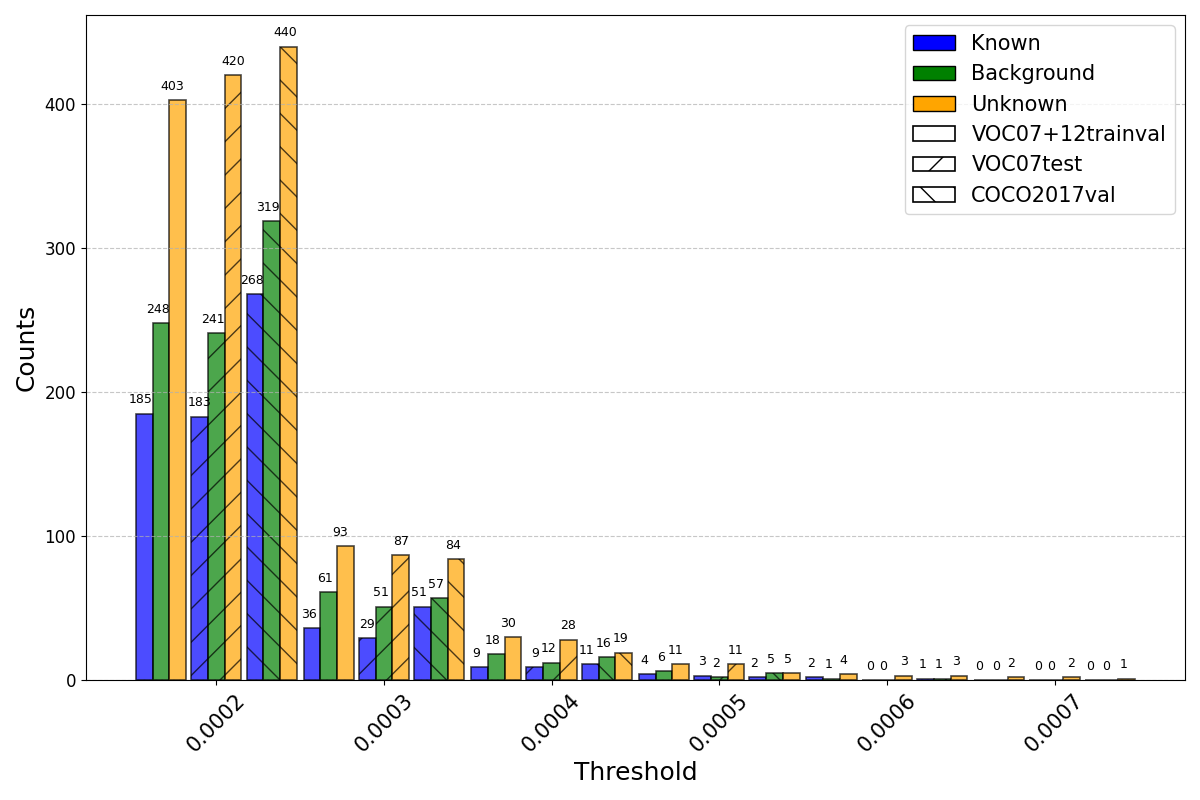}
		\caption{Frequency histogram of local features with different \(A_{local}\) threshold for three dataset settings.}
		\label{3datasets_agg}

\end{figure}

\textit{1) The analysis of aggregated gradients:} First, we conduct an in-depth study of the local aggregated gradient \(A_{local}\). We found that unknown classes have more abnormal gradient values, indicated by higher \(A_{local}\), compared to known and background classes. Additionally, as Fig. \ref{3datasets_agg} shown, we present the \(A_{local}\) values for 300 randomly selected proposals from known, background, and unknown classes across three datasets. The average results from three runs were shown as frequency histograms, with the x-axis representing different $A_{local}$ thresholds, which indicates the number of items greater than the threshold. We observed that as the threshold increases, the number of \(A_{local}\) occurrences for known and background classes approaches zero, while unknown classes consistently exhibit a certain number of $A_{local}$ values. This demonstrates that the unknown classes exhibit larger and more anomalously high \(A_{local}\) values. This aligns with our approach of selecting the top-\(k\) largest \(A_{global}\) as pseudo-unknown samples and the top-\(m\) largest \(A_{local}\) as abnormal local features. 

\begin{figure}
		\centering
		\includegraphics[width=0.46\textwidth]{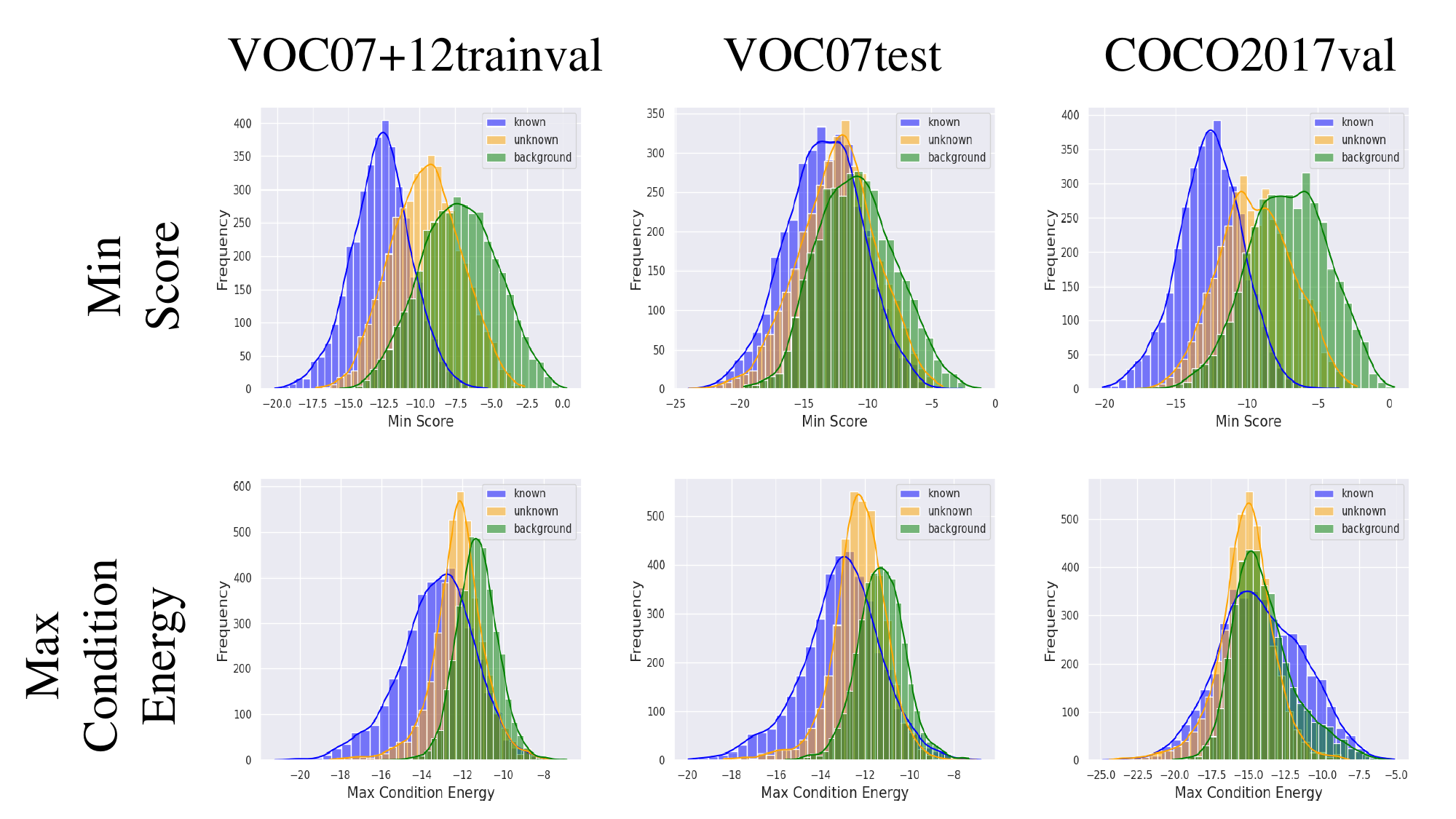}
		\caption{Statistics of the value distributions for different pseudo-unknown sample mining methods.}
		\label{psudo_diff}

\end{figure}

Then we delve into the global aggregated gradient \(A_{global}\), which serves as an uncertainty indicator for pseudo-unknown sampling. In Fig. \ref{psudo_diff}, we analyze the distribution of two other pseudo-unknown sample mining metrics \cite{ming2022delving,su2024toward} across known, background, and unknown classes, showing the three-peak distribution, including interference from the background class. Our method ensures that the background and known classes share the same distribution, validating the reasonableness of selecting the top-\(k\).
 Eq. (6) consists of two parts, we retained only the latter to form the ``ours-'' method. As shown in the first row of Fig. \ref{psudo_diff}, we found that the performance has already reached state-of-the-art levels. We added the non-zero attribution gradient count as an auxiliary term \cite{chen2024gaia} to form the complete version of \(A_{global}\), which minimizes the impact on known classes when training with pseudo-unknown samples and improves the accuracy of known classes.

\begin{figure}
        \centering
        \includegraphics[width=0.2\textwidth]{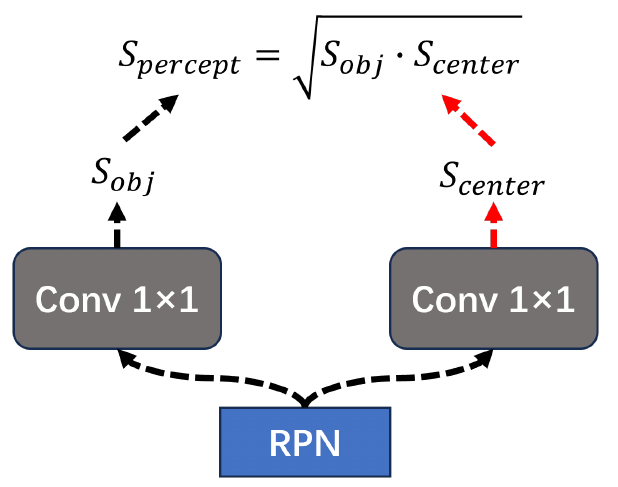}
        \caption{Our RPN structure, which attaches
a centerness branch parallel to the original objectness
branch.}
        \label{fig:RPN_struct}

\end{figure}

\begin{table}[!t]
        \centering
        \scriptsize
        \tabcolsep=0.15cm

        \begin{tabular}{cccc }
        \toprule
         {} & {\(S_{obj}\)} & {\(\sqrt{S_{obj} \cdot S_{center}}\)} \\
        \midrule
        VOC10-5-5 & 56.90 & \textbf{57.40} \\
        VOC-COCO & 38.10 & \textbf{40.20} \\
        \bottomrule
        \end{tabular}
                \caption{Average Recall with different RPNs}

        \label{tab:t4}
         \vspace{-1.5em}
\end{table}

\textit{2) The analysis of independently trained RPN:}
We utilize an independently trained backbone for the RPN and attach a centerness \cite{tian2020fcos} branch parallel to the original objectness branch (in Fig. \ref{fig:RPN_struct}), which can alleviate the issue of overfitting to known classes in the original RPN. As shown in Tab. \ref{tab:t4}, we conduct experiments on the choice of final object scores, which are trained only on the objectness branch and both two branches. The results show that for both VOC10-5-5 and VOC-COCO settings, our RPN structure and final object scores can perform better on average recall (\(AR\)) of objects.


\begin{figure} 
		\centering
		\includegraphics[width=0.40\textwidth]{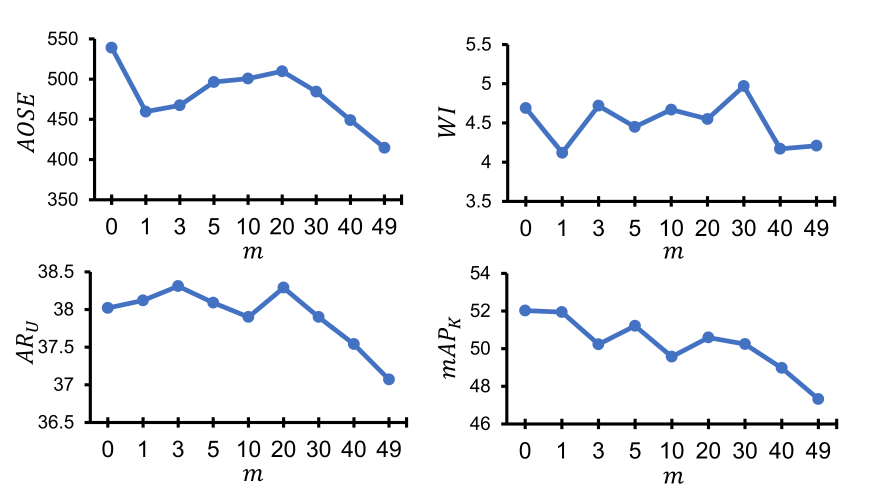}
		\caption{The choice of abnormal gradient feature number \(m\). We select \(m=1\) for all final results for better performance.}
		\label{abla_m}
\end{figure}


\subsection{Ablation study}

The following experiments are all conducted under the 1-shot VOC10-5-5 experimental setting with an average of 10 runs:

\textit{1) The abnormal gradient feature number \(m\):}  We ablate the abnormal feature number \(m\), as shown in Fig. \ref{abla_m}. The results indicate that including non-abnormal values in training compromises the precision of known classes and hinders the formation of effective unknown decision boundaries. Considering the best overall performance and additional computational overhead, we choose \(m=1\) by default.

\textit{2) The choice of \texorpdfstring {\(k\)}{} and \texorpdfstring {\(S_{fg:bg}\)}{}:}
We conduct ablation experiments on the number of pseudo-unknown sample mining \(k\) and the foreground-background mining ratio \(S_{fg:bg}\), as shown in Tab \ref{tab:t5}. When the ratio \(S_{fg:bg}\) remains constant, smaller values of \(k\) result in better \(WI\) and \(AOSE\) but poorer \(mAP_{K}\) and \(AR_{U}\). Conversely, larger values of \(k\) yield better known class accuracy \(mAP_{K}\) and \(AR_{U}\) but lower \(WI\) and \(AOSE\). We chose a balanced value of \(k=3\). When the mining number \(k\) remains constant, mining too few background proposals negatively affects all metrics. Therefore, we selected \(S_{fg:bg}=1:3\).

\begin{table}
        \centering
        \scriptsize

        \begin{tabular}{cc|ccccc }
        \toprule
         \(k\) & \(S_{fg:bg}\) & {\(WI\)} & {\(AOSE\)} & {\(mAP_{K}\)} & {\(R_{U}\)} & {\(AR_{U}\)} \\
        \midrule
        {3}&{1:3} & 4.12 & 459.60 & 51.94 & 79.88 & 38.12 \\
        {1}&{1:3} & 4.05 & \textbf{420.40} & 48.93 & 78.47 & 37.79 \\
        {5}&{1:3} & 4.58 & 501.17 & 50.80 & 79.18 & 37.96 \\
        {10}&{1:3} & 4.92 & 588.00 & \textbf{52.22} & \textbf{80.73} & \textbf{38.22} \\
        
        {3}&{1:1} & 4.35  & 486.00  & 50.42 & 79.44 & 38.29 \\
        {3}&{1:2} & \textbf{4.01}  & 439.00  & 50.87 & 79.33 & 37.80 \\
        {3}&{1:5} & 4.33  & 452.20  & 50.49 & 80.25 & 37.99 \\

        \bottomrule
        \end{tabular}
                \caption{Ablation study of \(k\) and \(S_{fg:bg}\)}

        \label{tab:t5}
        \end{table}

\begin{table}[!t]
        \centering

        \scriptsize

        \begin{tabular}{cc|ccccc }
        \toprule
         {} & {} & {\(WI\)} & {\(AOSE\)} & {\(mAP_{K}\)} & {\(R_{U}\)} & {\(AR_{U}\)} \\
        \midrule
        \multirow{2}{*}{1-shot} 
        & {CSC} & \textbf{4.12} & \textbf{459.60} & \textbf{51.94} & \textbf{79.88} & \textbf{38.12} \\
        {}&{UC} & 5.90 & 700.00 & 49.41 & 76.40 & 35.80 \\
        \midrule
        \multirow{2}{*}{3-shot} 
        & {CSC} & \textbf{3.72} & \textbf{451.20} & \textbf{53.09} & \textbf{80.55} & \textbf{39.53} \\
        {}&{UC} & 4.96 & 636.30 & 51.31 & 78.51 & 38.02 \\
        \midrule
        \multirow{2}{*}{5-shot} 
        & {CSC} & \textbf{3.78} & \textbf{512.20} & \textbf{54.35} & \textbf{81.37} & \textbf{40.32} \\
        {}&{UC} & 4.65 & 698.70 & 53.79 & 79.41 & 38.73 \\
        \midrule
        \multirow{2}{*}{10-shot} 
        & {CSC} & \textbf{3.43} & \textbf{546.30} & \textbf{58.55} & \textbf{79.39} & \textbf{39.79} \\
        {}&{UC} & 3.63 & 656.00 & 56.90 & 76.51 & 37.78 \\

        \bottomrule
        \end{tabular}
                \caption{Ablation study of prompt context type}
        \label{tab:t6}
\end{table}

\textit{3) The prompt context type:}
We conducted ablation experiments on the types of context used in prompt learning, specifically including \textbf{U}nified \textbf{C}ontext (UC) and \textbf{C}lass-\textbf{S}pecific \textbf{C}ontext (CSC). As shown in Tab. \ref{tab:t6}, we found that using CSC consistently outperforms UC. The main reason is that the object detection task generates diverse proposals, and using CSC can better capture the features of different classes.

\textit{4) The sensitivity analysis of \texorpdfstring {\(\lambda\)}{} and \texorpdfstring {\(\beta\)}{}:} We observed in Tab. \ref{lambda_analy} that as \(\lambda\) decreases, the \(AR_U\) gradually increases and stabilizes in a certain region. We chose the starting point of this phenomenon \(\lambda=0.0001\) for all experiments, as it provides balanced performance across other metrics as well. Tab. \ref{beta_analy} demonstrates that the variation in \(\beta\) has minimal impact on performance, therefore, we have chosen 1 as the default value for the weight factor in \(\boldsymbol{L}_{ADC}\).

\begin{table}
    \centering
    \scriptsize

    \begin{tabular}{l|ccccc}
        \toprule
        \(\lambda\) & {\(WI\)} & {\(AOSE\)} & {\(mAP_{K}\)} & {\(R_{U}\)} & {\(AR_{U}\)} \\
        \midrule
        0.1     & 4.49 & \textbf{356.40} & 49.77 & 63.20 & 32.50 \\
        0.01    & 4.49 & 436.90 & 51.37 & 70.21 & 35.25 \\
        0.001   & 4.61 & 464.50 & \textbf{52.17} & 75.09 & 37.00 \\
        0.0001  & \textbf{4.12} & 459.60 & 51.94 & 79.88 & \textbf{38.12} \\
        0.00001 & 4.16 & 418.80 & 50.64 & \textbf{80.62} & 37.38 \\
        0.000001& 4.18 & 411.90 & 50.00 & 80.41 & 36.13 \\
        \bottomrule
    \end{tabular}
    \caption{The sensitivity analysis of \(\lambda\)}
    \label{lambda_analy}
\end{table}

\begin{table}
    \centering
    \scriptsize

    \begin{tabular}{c|ccccc}
        \toprule
        \(\beta\) & {\(WI\)} & {\(AOSE\)} & {\(mAP_{K}\)} & {\(R_{U}\)} & {\(AR_{U}\)} \\
        \midrule
        0.1 & 4.22 & 467.80 & 51.86 & 79.58 & 37.90 \\
        0.3 & 4.29 & 456.40 & 50.74 & 79.48 & 37.52 \\
        0.5 & 4.24 & \textbf{443.60} & 51.28 & 78.98 & 37.70 \\
        0.7 & 4.58 & 473.10 & 50.50 & 79.13 & 37.50 \\
        1 & \textbf{4.12} & 459.60 & \textbf{51.94} & \textbf{79.88} & \textbf{38.12} \\
        \bottomrule
    \end{tabular}
        \caption{The sensitivity analysis of \(\beta\)}
    \label{beta_analy}
    \vspace{-1em}
\end{table}

\begin{figure*}[!t]
\centering
\includegraphics[width=0.88\textwidth]{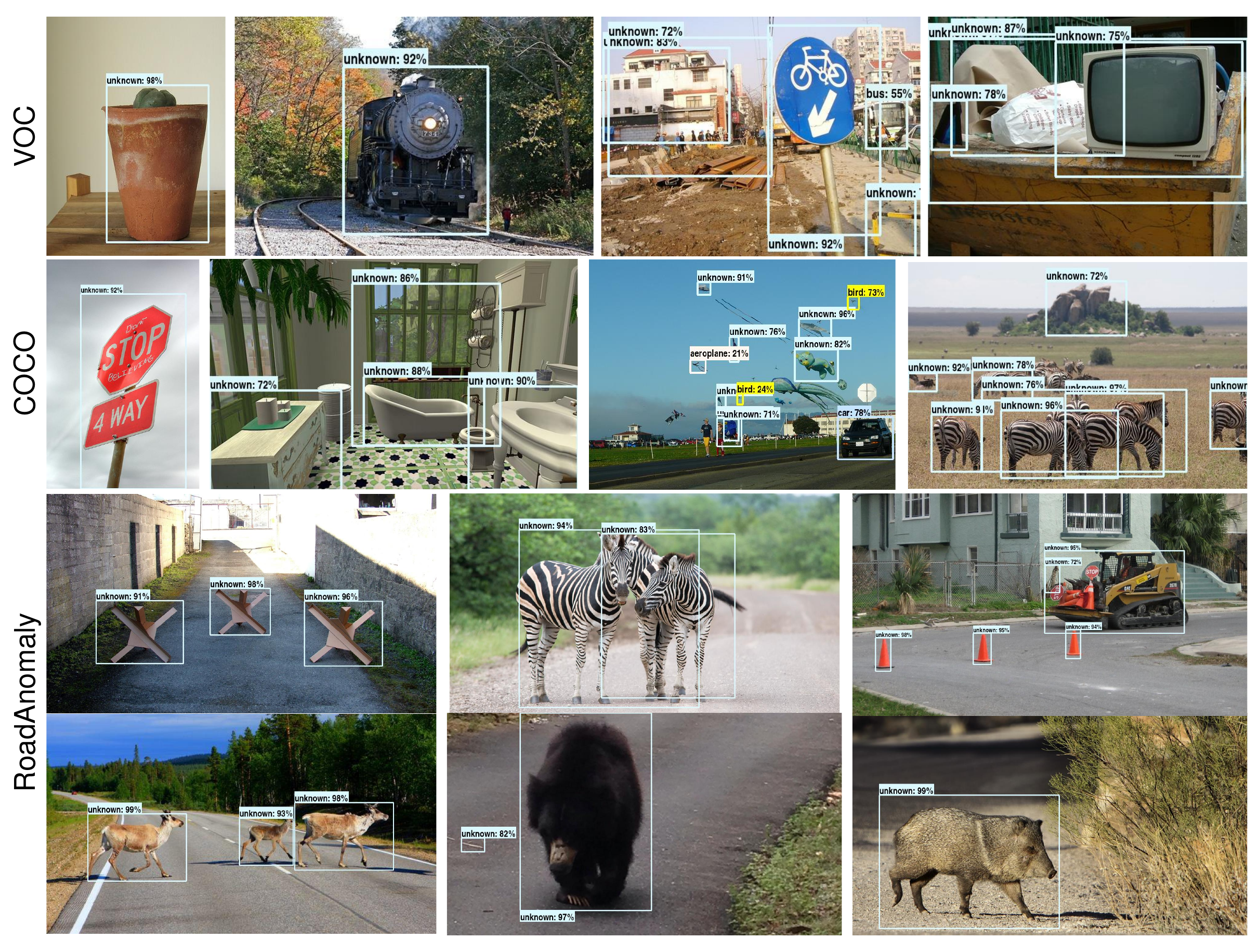}

\caption{More visualization results on VOC, COCO, and RoadAnomaly datasets under 10-shot VOC10-5-5 setting.}

\label{fig:appendix_vis}
\end{figure*}

\subsection{More visualization results}
Fig. \ref{fig:appendix_vis} presents additional visualization results of our approach on the VOC, COCO, and RoadAnomaly datasets. It can be observed that our method demonstrates a capacity to perceive numerous objects present in the images and establish decision boundaries effectively for their classification.

\end{document}